\documentclass[runningheads]{llncs}
\usepackage{cite}
\usepackage{amsmath,amssymb,amsfonts}
\usepackage{algorithmic}
\usepackage{graphicx}
\usepackage{textcomp}
\usepackage{xcolor}
\usepackage[switch]{lineno}
\usepackage{fancyhdr}
\usepackage{bm}
\usepackage{multirow}
\usepackage{booktabs}
\usepackage{float}
\usepackage[inkscapelatex=false]{svg} 
\usepackage{epstopdf}
\usepackage[linesnumbered,ruled,vlined]{algorithm2e}
\usepackage{marvosym}
\usepackage{subfiles}
\usepackage{appendix}

\newcommand{\labinfo}{\thanks{Hanbing Liu, Jingge Wang, Xuan Zhang, and Yang Li are from the Shenzhen Key Laboratory of Ubiquitous Data Enabling, SIGS, Tsinghua University.}}

\begin{document}

\title{Enhancing Continuous Domain Adaptation with Multi-Path Transfer Curriculum}
\titlerunning{W-MPOT}
%
\author{Hanbing Liu\orcidID{0000-0002-5988-2306} \and
Jingge Wang\orcidID{0000-0002-9231-882X} \and
Xuan Zhang\orcidID{0000-0002-7537-4002
} \and
Ye Guo\orcidID{0000-0002-5268-5289} \and
Yang Li\textsuperscript{(\Letter)}\orcidID{0000-0002-2053-6393}}
\authorrunning{Hanbing Liu et al.}
%
\institute{Tsinghua Shenzhen International Graudate School (SIGS), Tsinghua University\labinfo \\
\email{\{liuhb21,wjg22\}@mails.tsinghua.edu.cn} \\
\email{\{xuanzhang,guo-ye,yangli\}@sz.tsinghua.edu.cn}}


%
\maketitle              
\begin{abstract}
Addressing the large distribution gap between training and testing data has long been a challenge in machine learning, giving rise to fields such as transfer learning and domain adaptation. Recently, Continuous Domain Adaptation (CDA) has emerged as an effective technique, closing this gap by utilizing a series of intermediate domains. 
This paper contributes a novel CDA method, W-MPOT, which rigorously addresses the domain ordering and error accumulation problems overlooked by previous studies. Specifically, we construct a transfer curriculum over the source and intermediate domains based on Wasserstein distance, motivated by theoretical analysis of CDA. Then we transfer the source model to the target domain through multiple valid paths in the curriculum using a modified version of continuous optimal transport. A bidirectional path consistency constraint is introduced to mitigate the impact of accumulated mapping errors during continuous transfer. We extensively evaluate W-MPOT on multiple datasets, achieving up to 54.1\% accuracy improvement on multi-session Alzheimer MR image classification and 94.7\% MSE reduction on battery capacity estimation.

\keywords{Continuous Domain Adaptation \and Wasserstein distance \and Transfer curriculum \and Optimal Transport \and Path Consistency regularization.}
\end{abstract}

\section{Introduction}

Domain shift is a common challenge in many real life applications \cite{wang2023empirical}. For example, in medical imaging, models trained on the data from one institution may not generalize well to another institution with different imaging hardware. Similarly, in battery capacity monitoring, models trained on lab-collected data may perform poorly under diverse operation environments. Obtaining the annotation for the new domains, however, is often very costly or infeasible. To address this challenge, Unsupervised Domain Adaptation (UDA) has been proposed, leveraging the labeled data from the source domain to improve the performance of learning models on the unlabeled target domain \cite{ganin2015unsupervised}. In particular, UDA aims to align the distributions of the source and target domains using labeled source data and unlabeled target data, typically by learning domain-invariant representations \cite{sun2017correlation} or adversarial learning schemes. Nevertheless, one challenge of UDA is its limited effectiveness when confronted with significant domain shift. Studies conducted by Zhao et al. \cite{zhao2019learning} have shed light on the relationship between the domain shift and generalization error in UDA. They have shown that the effectiveness of UDA is bounded by the distributional divergence between the source and target domain, so the performance of the adapted model on the target domain may not be satisfactory with a substantial domain shift. In addressing this challenge, many works studied the problem of Continuous Domain Adaptation (CDA) \cite{xu2022delving}.

Instead of directly adapting the model from the source to the target domain, CDA captures the underlying domain continuity leveraging a stream of observed intermediate domains, and gradually bridges the substantial domain gap by adapting the model progressively. 
There are various applications of CDA requiring continuous domains with indexed metadata \cite{liu2020learning, hoffman2014continuous}. For example, in medical data analysis, age acts as a continuous metadata for disease diagnosis across patients of different age groups. In the online battery capacity estimation problem, the state of health (SoH) of the battery acts as a continuous metadata that differs across batteries. CDA has attracted a great deal of attention and gained a rapid performance boost by self-training \cite{kumar2020understanding, zhou2022active}, pseudo-labeling \cite{liang2020we}, adversarial algorithms, optimal transport (OT) \cite{ortiz2019cdot} and so on. In particular, Oritiz et al. \cite{ortiz2019cdot} designed an efficient forward-backward splitting optimization algorithm for continuous optimal transport (COT) and demonstrated the efficacy of OT in reducing the domain shift and improving the performance of adaptation models.

While there have been significant advances in CDA, it still faces two critical challenges, determining the transfer order of intermediate domains in the absence of continuous metadata and mitigating cumulative errors throughout the continuous adaptation process. For the first issue, metadata could be missing or incorrect, and sometimes metadata alone can not fully explain the difference between data distributions. The proper ordering of intermediate domains is significant for CDA in transferring knowledge all the way to the target domain. Indeed, it is necessary to order the intermediate domains to facilitate continuous transfer without relying on explicit metadata. As a divergence measurement that takes into account the geometry of the data distributions, Wasserstein distance (w-distance) \cite{villani2009wasserstein} plays an important role in deriving the generalization bound in domain adaptation, which implies the effectiveness of reducing domain divergence in the Wasserstein space.
In this work, we propose a transfer curriculum in the Wasserstein space, aimed at determining the optimal sequence of intermediate domains for better knowledge transfer. 
The incorporation of Wasserstein-based transfer curriculum provides a principled and effective way to order the intermediate domains, enabling more precise and controlled knowledge transfer. 
A more comprehensive discussion regarding the process of selecting the appropriate transfer sequence, which ultimately leads to a tighter generalization bound, will be provided in the method section. 

For the second issue, as the model progressively adapts to new domains, errors can accumulate and degrade the overall performance. The accumulation of errors can occur during CDA due to the successive estimation of pseudo-labels or intermediate adaptation results, e.g., the error accumulates during each projection of source domain data based on the estimated optimal transport map in \cite{ortiz2019cdot}. Fourier domain filtering possesses the capability to mitigate cumulative errors \cite{durak2010adaptive}, but its efficacy is limited to fixed frequencies of errors, thereby exhibiting inadequacies in terms of flexibility and robustness. To tackle this challenge, we introduce a path consistency regularization scheme for OT-based CDA inspired by \cite{ortiz2019cdot}. Our multi-path regularization scheme enforces consistency among multiple transfer paths, effectively reducing the impact of accumulated errors and improving the robustness and stability of the transferred model.

In summary, the main contribution of this work lies in four aspects:

\begin{itemize}
\item[1)] \textbf{W-MPOT}: The paper proposes a novel CDA framework, named W-MPOT, which incorporates a Wasserstein-based transfer curriculum and multi-path consistency regularization, providing a principled and effective solution for CDA in scenarios where explicit metadata is unavailable.

\item[2)] \textbf{Wasserstein-based Transfer Curriculum}: 
The method employs w-distance to devise the transfer curriculum, providing theoretical proofs and generalization upper bounds on the error incurred by improper sorting based on w-distance.

\item[3)] \textbf{Multi-Path Optimal Transport}: The paper introduces a multi-path domain adaptation method based on Optimal Transport, namely MPOT, to enforce consistency among multiple adaptation paths. By mitigating the impact of accumulated errors during continuous transfer, MPOT significantly enhances the overall performance and stability of the adaptation process.

\item[4)] \textbf{Comprehensive Empirical Validation}: We conduct a thorough set of experiments to validate the motivation and effectiveness of our proposed methods. These experiments cover various domains, including \textit{ADNI, Battery Charging-discharging Capacity and Rotated MNIST} datasets, and demonstrate the superiority of our approach compared to alternative methods. 

\end{itemize}

\begin{figure}[h]
	\begin{center}
		\centerline{\includegraphics[width=0.9\columnwidth]{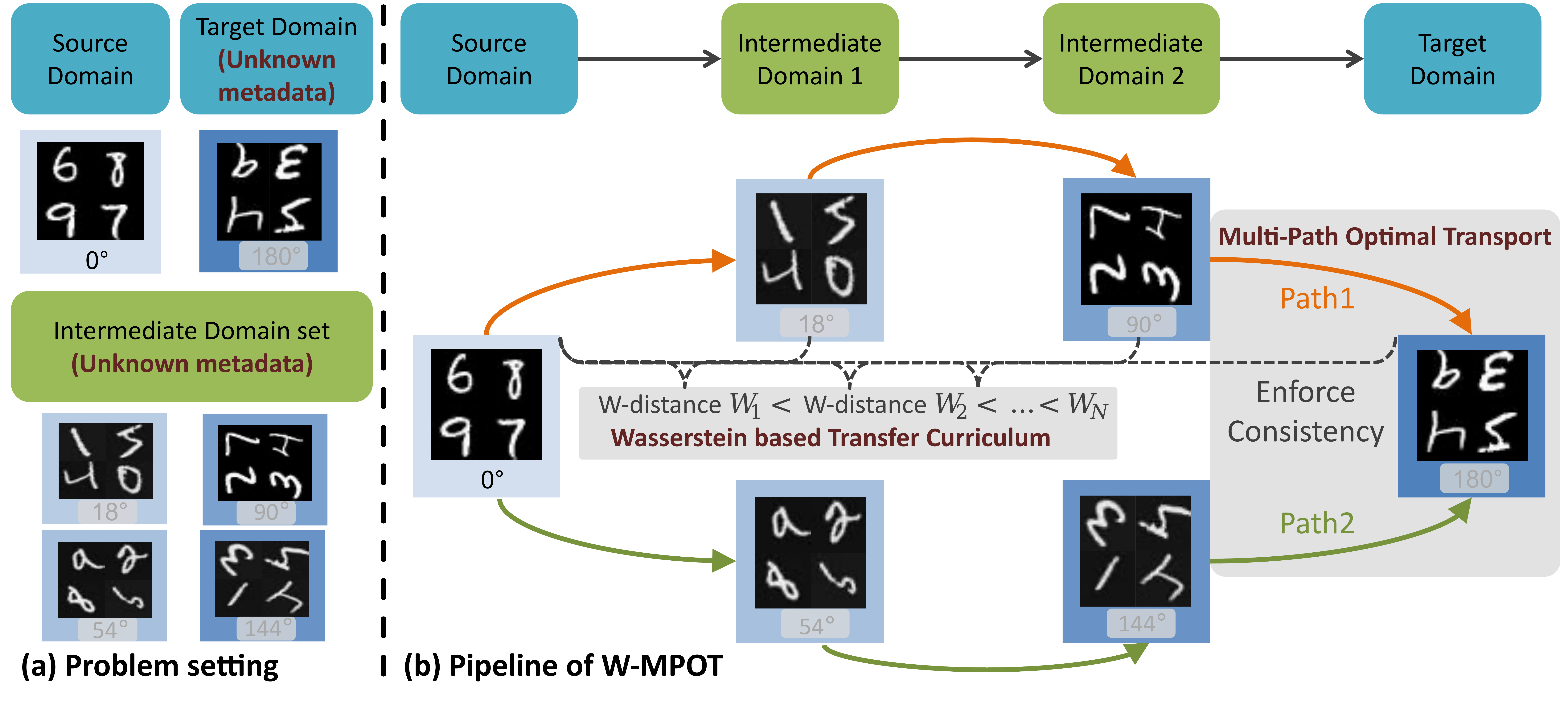}}
		\caption{\textbf{Illustration of our proposed W-MPOT.} (a) In the \textit{Rotated MNIST} example, we are given a source domain of 0 degrees and a target/intermediate domain set with unknown degrees. (b) In our proposed W-MPOT, we address the challenge of unknown domain metadata (angles) and perform CDA. We utilize Wasserstein based transfer curriculum to sort intermediate domains and employ MPOT to enforce consistency, thereby enhancing the transfer effectiveness.}
    \label{fig:intro}
	\end{center}
 \vspace{-1cm}
\end{figure}

\section{Methodology}
\label{sec:methodology}

\subsection{Preliminary}
We employ optimal transport (OT) to map the source domain into the target domain. OT provides a measurement of the divergence between two probability distributions by finding an optimal transportation plan that minimizes the total cost of mapping one distribution to the other \cite{courty2016optimal}. Detailed explanations regarding OT can be found in Supplementary A.

Consider the scenario where a labeled source domain and a collection of unlabeled auxiliary domains are available.
Let $\mathcal{X}\subset \mathbb{R}^d$ be the feature space and 
{$\mathcal{Y}\subset \mathbb{R}$} be the label space. 
Denote the source domain and target domain as $D_S=\left\{\left(\boldsymbol{x}_{\boldsymbol{j}}, y_j\right)\right\}_{j=1}^{N_S}$, $D_T=\left\{\boldsymbol{x}_{\boldsymbol{j}}\right\}_{j=1}^{N_T}$, where  $N_S$ and $N_T$ are the number of samples in the source domain and target domain, respectively.
The candidate set of intermediate domains is detonated as $\mathcal{D}_{I}=\left\{D_{I_1}, D_{I_2}, \ldots, D_{I_K}\right\}$, where each domain is denoted as $D_{I_k}=\left\{\boldsymbol{x}_{\boldsymbol{j}}\right\}_{j=1}^{N_{I_k}}, k = 1, \ldots, K$ and $N_{I_k}$ denotes the number of unlabeled data.
Let $\mu_S$, $\mu_{I_k}$, $\mu_T\in \mathcal{P}(\mathcal{X})$ be the probability measures on $\mathbb{R}^d$ for the source, intermediate, and target domains, respectively.
The objective is to make predictions $\left\{\widehat{y}_j\right\}_{j=1}^{N_T}$ for samples in the target domain.

While the order of intermediate domains in CDA can be determined using available metadata, the challenge arises when the metadata is absent, requiring further efforts to determine the proper order for CDA.
Given the $K$ candidate intermediate domains, a transfer curriculum defines an ordered sequence of $N$ domains from $\mathcal{D}_I$, denoted as $\widehat{\mathcal{D}}_I$, to be used as intermediate domains for the CDA.

\subsection{Method Framework}
The framework of our proposed W-MPOT, comprising the Wasserstein-based transfer curriculum module and the Multi-Path Optimal Transport (MPOT) module, is depicted in Fig. \ref{fig:intro}. The Wasserstein-based transfer curriculum module selects an optimal intermediate domain sequence $\widehat{\mathcal{D}}_I$ for a given target domain $D_T$. Specifically, after calculating the w-distance \cite{villani2009wasserstein} between each intermediate domain and the source domain, 
we sort intermediate domains closer to the source than the target domain by their w-distances. 
In the MPOT module, we adopt COT \cite{ortiz2019cdot} to transfer the source domain knowledge through 
the sorted intermediate domain sequence, and finally adapt to the target domain in an unsupervised manner. A multi-path consistency term is proposed to regularize the continuous adaptation. Since the divergence between the source and target domains is minimized by the OT-based adaptation process, the final prediction for the target domain can be derived by a regressor or classifier trained on the transported source domain.

\subsection{Wasserstein-based Transfer Curriculum}

Properly ordering the intermediate domains ensures a more effective transfer of knowledge. Here we present the motivation for arranging intermediate domains in the Wasserstein space with a simple example.

Given the labeled source domain $D_S$ , the target domain $D_T$ and two candidate intermediate domains  $D_{I_1}$ and $D_{I_2}$.
We assume that the optimal transfer order is $D_S \rightarrow D_{I_1} \rightarrow  D_{I_2} \rightarrow D_T$, i.e., the intermediate domain $D_{I_1}$ is closer to $D_{S}$ than to $D_{T_2}$. 
Apparently, another possible transfer order would be $D_S \rightarrow D_{I_2} \rightarrow  D_{I_1} \rightarrow D_T$. 
We first present the generalization bound proposed by \cite{shen2018wasserstein} which relates the source and target errors using w-distance. 
Let $\mu_{I_1}$, $\mu_{I_2}\in \mathcal{P}(\mathcal{X})$ be the probability measures for domains $D_{I_1}$, $D_{I_2}$, respectively. $h$ and $f$ denote the predicted hypothesis and the true labeling function, respectively. $\epsilon_{\mu}(h,f)=\mathbb{E}_{\boldsymbol{x}\sim\mu}\left[\left|h(\boldsymbol{x}) - f(\boldsymbol{x})\right|\right]$ and $\epsilon$ is the combined error of the ideal hypothesis $h^*$. Assume that all hypotheses in the hypothesis set $H$ satisfies the $A$-Lipschitz continuous condition for some $A$. 

By simply applying the lemma proposed in \cite{shen2018wasserstein} to each source-target pairs $(D_S, D_{I_1})$, $(D_{I_1}, D_{I_2})$ and $(D_{I_2}, D_{T})$, the generalization bound of the transfer path $D_S \rightarrow D_{I_1} \rightarrow  D_{I_2} \rightarrow D_T$ could be derived as the following equation. For a detailed explanation of the lemma, please refer to Supplementary B.
\begin{equation}
\begin{split}
\epsilon_{\mu_T}(h,f) &\leq \epsilon_{\mu_S}(h,f)
+2 A \cdot\mathcal{W}_{1}\left(\mu_{S}, \mu_{I_1}\right) \\
&+2 A \cdot\mathcal{W}_{1}\left(\mu_{I_1}, \mu_{I_2}\right) 
 +2 A \cdot\mathcal{W}_{1}\left(\mu_{I_2}, \mu_{T}\right)
+\epsilon,
\end{split}
\end{equation}

Similarly, for another transfer path $D_S \rightarrow D_{I_2} \rightarrow  D_{I_1} \rightarrow D_T$, the generalization bound is
\begin{equation}
\begin{split}
\epsilon_{\mu_T}(h,f) &\leq \epsilon_{\mu_S}(h,f)
+2 A \cdot\mathcal{W}_{1}\left(\mu_{S}, \mu_{I_2}\right) \\
&+2 A \cdot\mathcal{W}_{1}\left(\mu_{I_2}, \mu_{I_1}\right) 
 +2 A \cdot\mathcal{W}_{1}\left(\mu_{I_1}, \mu_{T}\right)
+\epsilon,
\end{split}
\end{equation}
where $\epsilon$ is the same. According to the optimal transfer path assumption, it is straightforward to get $\mathcal{W}_{1}\left(\mu_{S}, \mu_{I_2}\right) > \mathcal{W}_{1}\left(\mu_{S}, \mu_{I_1}\right)$, and $\mathcal{W}_{1}\left(\mu_{I_1}, \mu_{T}\right) > \mathcal{W}_{1}\left(\mu_{I_2}, \mu_{T}\right)$. 
Therefore, better domain transfer order $D_S \rightarrow D_{I_1} \rightarrow  D_{I_2} \rightarrow D_T$ will lead to tighter generalization bound. The optimal transfer path order will achieve better performance on the target domain than the other, which justifies the use of the w-distance in our transfer curriculum.

This result leads to our Wasserstein-based transfer curriculum to select and sort for intermediate domains. We use the following form of w-distance to measure the closeness between each intermediate domain $D_{I_k}$ and the source domain $D_S$,
\begin{equation}
\begin{gathered}
  W_k=\min _\gamma\langle\gamma_k, \mathbf{M}_k\rangle_F+\lambda \cdot \Omega(\gamma_k)  \\
  \text { s.t. } \gamma_k \mathbf{1}  = \mu_S \quad
\gamma_k^T \mathbf{1}  = \mu_{I_k}, k = 1, \ldots, K,
\end{gathered}
\end{equation}

\noindent where $\gamma_k \geq 0$ is the transport matrix and $\mathbf{M}_k$ is the cost matrix between the distribution $\mu_S$ and $\mu_{I_k}$. 
$W_k$ measures the similarity between each intermediate domain and the source domain, i.e., the greater the $W_k$ is, the farther the intermediate domain is from the source domain. Note that any intermediate domain that is further from the source domain than the target domain is discarded. The remaining $N$ domains in the intermediate domain set are then sorted in order of $W_k$ and we could obtain a domain sequence $\widehat{\mathcal{D}}_{I}=  \widehat{D}_{I_1}\rightarrow \widehat{D}_{I_2}\rightarrow \ldots \rightarrow \widehat{D}_{I_N}$, where $W_1 \leq W_2 \ldots \leq W_N$.
As a result, the Wasserstein-based transfer curriculum generates a sorted transfer sequence $\widehat{\mathcal{D}}_{I}$ that represents the desired order of the domains, arranged from those closer to the source domain to those farther away.
By utilizing the w-distance to sort multiple intermediate domains, we eliminate the need for meta-information.


\subsection{Multi-Path Optimal Transport}

We apply OT-based domain adaptation iteratively across each intermediate domain indexed by the proposed curriculum. In order to make predictions for the target domain $D_T$, MPOT constructs sequential transport plans, denoted as $\gamma_m$, through a series of successive steps from the source to the target domain. The steps are as follows. Given the sorted sequence of intermediate domain  $\widehat{\mathcal{D}}_{I}$, the source domain is initially mapped to the first intermediate domain $\widehat{D}_{I_1}$ using direct OT \cite{courty2016optimal} and we can derive the first transport plan $\gamma_0$. Following this, the barycentric mapping of the source domain to the first intermediate domain $\widehat{D}_{I_1}$ can be defined by a weighted barycenter of its neighbors, $\mathcal{B}_{I_1}(D_S)=N_S \cdot \gamma_0 \cdot \widehat{D}_{I_1}$. The distribution of the mapped source domain on $\widehat{D}_{I_1}$ is denoted as $\mu_{SI_{1}}$, which is consistent with the distribution of the target intermediate domain $\mu_{I_1}$.

Then for the following intermediate domains, $\widehat{D}_{I_n}, n \in 2, \ldots, N$, the probabilistic coupling $\gamma_{n-1}$ between the domain $\widehat{D}_{I_{n-1}}$ and the subsequent domain $\widehat{D}_{I_n}$ is calculated using COT \cite{ortiz2019cdot},

\begin{equation}
\begin{gathered}
\gamma_{n-1}=\underset{\gamma \in \mathbb{R}^{N_{S} \times N_{T}} }{\operatorname{argmin}}\left\langle\gamma, \mathbf{M}^{[n-1, n]}\right\rangle+\lambda \Omega(\gamma)+\eta_t R_t(\gamma) \\
\text { s.t. } \gamma \mathbf{1}  = \mu_{SI_{n-1}} \quad
\gamma^T \mathbf{1}  = \mu_{I_n} \quad
\gamma  \geq 0, \\
\end{gathered}
\label{eq:cot}
\end{equation}
where $\mathbf{M}^{[n-1, n]}$ is the cost matrix defining the cost to move mass from the distribution of $\mu_{SI_{n-1}}$ to $\mu_{I_{n}}$. $\Omega(\cdot)$ denotes the entropic regularization term $\Omega(\gamma)=\sum_{i, j} \gamma_{i, j} \log \left(\gamma_{i, j}\right)$, and $\lambda > 0$ is the weight of entropic regularization. $R_t(\cdot)$ is a time regularizer with coefficients $\eta_t >0$, which aims to enforce smoothness and coherence across consecutive time steps \cite{ortiz2019cdot}.

\begin{figure}[h]
	\begin{center}
		\centerline{\includegraphics[width=0.9\columnwidth]{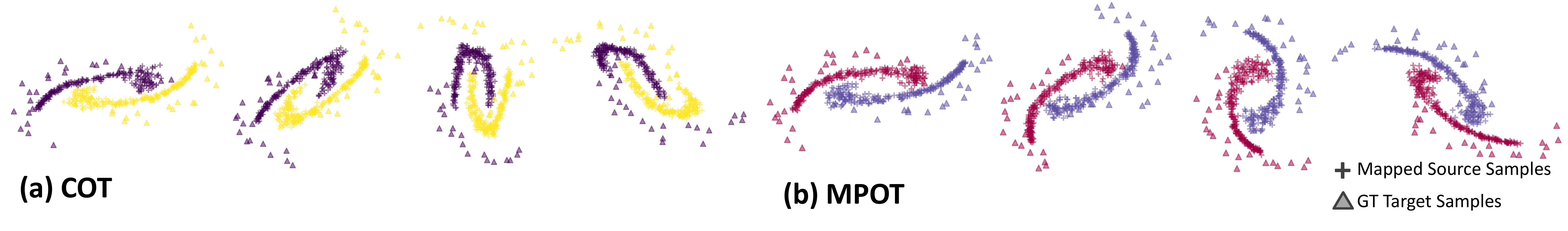}}
		\caption{\textbf{The Effect of Multi-Path regularization on the Optimal Transport of  source domain sample.} Visual experiments are conducted on simulated half-moon data to compare the migration effects of (a) COT with our proposed (b) MPOT. The source domain has an angle of 0 degrees, and the angles on the graph increase by 36 degrees from left to right. (a) and (b) depict the mapping results obtained by applying the COT and MPOT methods respectively to sequentially map the source domain to the target domains. The triangles represent the ground truth target domain data, while the plus signs represent the mapped source domain data after the adaptation.}
    \label{fig:map}
	\end{center}
 \vspace{-1cm}
\end{figure}

Upon the completion of the sequential transfer from the source domain to all intermediate domains, the transport plan $\gamma_{N}$ from $\widehat{D}_{I_N}$ to the target domain $D_T$ will be computed using MPOT with the equation as follows,
\begin{equation}
\begin{split}
\gamma_{N}=\underset{\gamma \in \mathbb{R}^{N_{S} \times N_{T}} }{\operatorname{argmin}}&\left\langle\gamma, \mathbf{M}^{[N, N+1]}\right\rangle+\lambda \Omega(\gamma) +\eta_t R_t(\gamma) + \eta_p R_p(\gamma, \gamma_{p_2}) \\
\text { s.t. } \gamma \mathbf{1}  = &\mu_{SI_{N}} \quad
\gamma^T \mathbf{1}  = \mu_{T} \quad
\gamma  \geq 0, \\
\end{split}
\label{eq:mpot}
\end{equation}
where $\mathbf{M}^{[N, N+1]}$ is the cost matrix between the distribution $\mu_{SI_{N}}$ and $\mu_{T}$. $\eta_t >0$ and $\eta_p>0$ are the coefficients to adjust the weights. We further introduce a path consistency regularizer $R_p(\cdot)$ by comparing it with another transfer path,
\begin{equation}
\begin{split}
R_p(\gamma, \gamma_{p_2})=\left\|N_S \cdot \gamma \cdot \widehat{D}_{I_n}-N_S \cdot \gamma_{p_2} \cdot \widehat{D}_{I_{n}}\right\|_F^2,
\end{split}
\label{eq:rp_f}
\end{equation}
where $\gamma_{p_2}$ is the transport plan of the second possible path which is utilized to refine the $\gamma$ of Path 1. During each adaptation step across intermediate domains in the curriculum, the incremental steps introduce minor inaccuracies, potentially impacting overall transfer performance over time. By employing $R_p(\cdot)$, we exploit the complementary information present in diverse paths to alleviate the accumulation of errors. The continuous adaptation results of the original COT and our proposed MPOT on the simulated half-moon dataset are shown in Fig.\ref{fig:map}.

During continuous adaptation, the COT method experiences a substantial increase in cumulative error, leading to deteriorating adaptation results as the rotation angle increases. In contrast, our proposed method, MPOT, effectively addresses the issue of cumulative errors by leveraging multiple transfer paths. 

Notablely, the paths interact in a bidirectional manner within $R_p(\cdot)$: not only does Path 2 impose constraints on Path 1, but Path 1 likewise restricts Path 2. This creates a reciprocal dynamic and both paths can be jointly optimized simultaneously. Moreover, the above OT problem (\ref{eq:mpot}) fits into the forward-backward splitting algorithm \cite{bui2021bregman} as in \cite{ortiz2019cdot}, whose solution could be efficiently computed with the Sinkhorn algorithm \cite{cuturi2013sinkhorn}. More details about the solution for the Sinkhorn algorithm are given in Supplementary C. Let $\mathcal{J}(\cdot)$ be the combination of regularization terms. The details of the bidirectional optimization procedure are depicted in the subsequent Algorithm \ref{algr:training}.

\begin{algorithm}[ht]
\DontPrintSemicolon
  \KwIn{Transport matrix of Path 1 $\gamma_{p_1}$, Transport matrix of Path 2 $\gamma_{p_2}$, step size $\alpha$, cost matrix of Path 1 $\mathbf{M}^{[1]}$, cost matrix of Path 2 $\mathbf{M}^{[2]}$, weight of Path 1 $\lambda_1$ and Path 2 $\lambda_2$, iteration times $c$}
  \KwOut{Refined transport matrix from $N$-th intermediate domain to target domain $\gamma_{N}'$}
  
  \BlankLine

\textbf{Initialize: } $\gamma_0 \in (0, +\infty)^{N_S \times N_T}$

\For{$c \leftarrow 0, 1, ...$}{

$\mathbf{M}_c^{[1]}=\alpha \mathbf{M}^{[N, N+1]} + \alpha \nabla \mathcal{J} (\gamma_c, \gamma_{p_2})$ \\
$\mathbf{M}_c^{[2]} = \alpha \mathbf{M}^{[N, N+1]} + \alpha \nabla \mathcal{J} (\gamma_c, \gamma_{p_1})$ \\
$\mathbf{M}_c = \lambda_1 \mathbf{M}_c^{[1]} + \lambda_2 \mathbf{M}_c^{[2]}$ \\
$\gamma_{c+1} = \operatorname{Sinkhorn}\left(\mathbf{M}_c, 1+\alpha \lambda, \mu_{SI_{N}}, \mu_{T}\right) $ \\
}

$\gamma_{N}' = \gamma_\infty$

\caption{Bidirectional Optimization algorithm in MPOT}
\label{algr:training}
\end{algorithm}

Once the refined transport matrix, denoted as $\gamma_{N}'$, has been computed, the barycentric mapping from the mapped source domain to the target domain can be obtained by using $\mathcal{B}_T(D_S^{I_N})=N_S \cdot \gamma_{N}'\cdot D_T$, where $D_S^{I_j}$ denotes the domain mapped from the source domain $D_S$ to $\widehat{D}_{I_j}$. Consequently, a classifier or regressor can be trained on $\mathcal{B}_T(D_S^{I_N})$ using the available source labels and can be directly deployed on the target domain for various applications.


\section{Experimental Results}
\label{sec:experiment}

\subsection{Datasets and Experimental Configurations}

Experiments are conducted on three datasets, each offering unique characteristics and challenges. Details regarding the implementation are presented in Supplementary D.

\textbf{ADNI.} The Alzheimer's Disease Neuroimaging Initiative (ADNI) dataset is a crucial resource for Alzheimer's disease research, housing a repository of MRI images that furnish in-depth structural insights into the brain \cite{petersen2010alzheimer}. The 3D MRI data is sliced along the direction of the vertical spinal column, resulting in 2D MRI images. The 2D MRI images are categorized into a source domain (ages 50-70, 190 samples) and intermediate domains (ages 70-72, 72-74, 74-76, 76-78, 78-82, 82-92, 50 samples each). The primary task is classifying MRI images into five disease categories. In our ADNI MRI image experiments ($128\times128$ dimensions), we performed image normalization to a 0-1 range. Subsequently, we utilized a pre-trained VGG16 model, originally trained on the ImageNet dataset, to extract image features. The VGG16 model's output consists of a $512\times4\times4$ feature map, which is condensed into a 16-dimensional feature vector by averaging across the channel dimension.

\textbf{Battery Charging-discharging Capacity.} The capacity of lithium-ion batteries holds paramount significance in the context of power systems and electric vehicles. Laboratory experiments are conducted via a pulse test \cite{zhou2020fast} to collect voltage-capacity data pairs on batteries during charging and discharging processes at different state of charge (SoC) levels. The dataset includes a source domain (5\% SoC) and nine intermediate domains (10\%-50\% SoC), each with 67 samples. It is a regression problem evaluated with Mean Squared Error (MSE).

\textbf{Rotated MNIST.} The Rotated MNIST dataset is a variation of the MNIST dataset, featuring rotated digit images. It is divided into five domains with different rotation angles. The source domain (0 degrees) and intermediate domains (18, 36, 54, 72, and 90 degrees) each contain 1000 samples.

\subsection{Analysis of Wasserstein-based Transfer Curriculum}
This section assesses the effectiveness of our proposed Wasserstein-based transfer curriculum, comparing it with two alternative domain adaptation methods. We examine Directly Optimal Transport (DOT), which transfers data directly from the source to the target domain, and Continuously Optimal Transport (COT), a progressive adaptation approach. COT is explored in two configurations: COT+Metadata, utilizing genuine domain metadata for sorting intermediate domains, and COT+W-dis, employing our Wasserstein-based transfer curriculum in the absence of metadata. In our experimental design, we maintain consistency between the source and target domains across all methods.  

\begin{figure}[h]
	\begin{center}
		\centerline{\includegraphics[width=0.9\columnwidth]{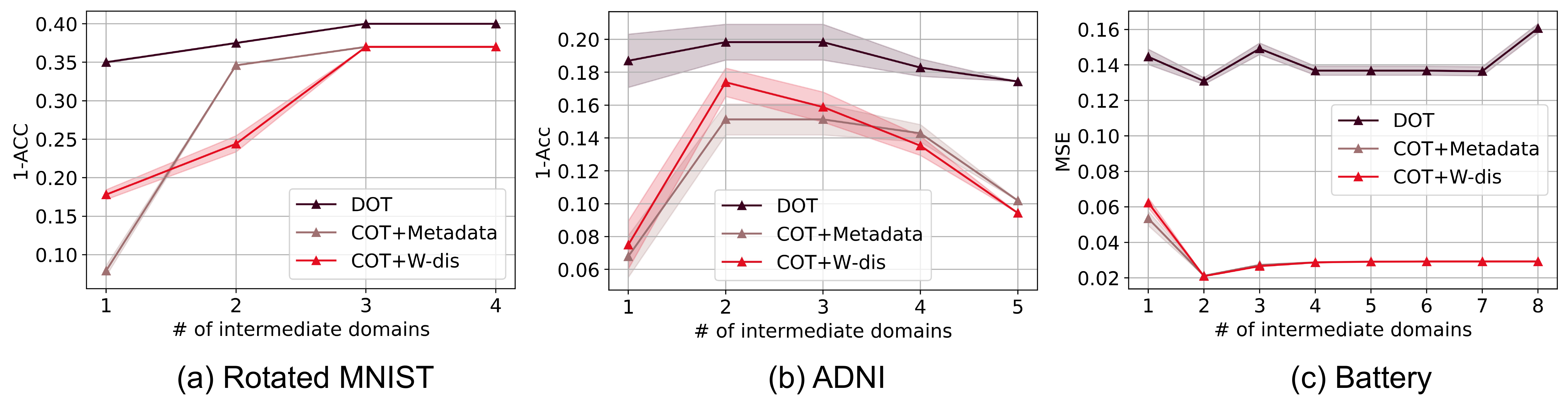}}
		\caption{\textbf{Domain ordering Results.} The results for the (a)Rotated MNIST, (b)ADNI, and (c)Battery Charging-discharging Capacity datasets are shown in this figure. The evaluation metric for Rotated MNIST and ADNI is accuracy, while for the Battery dataset is MSE. The methods compared are DOT (Direct Optimal Transport) and COT (Continuous Optimal Transport). Two sorting approaches are utilized: metadata-based sorting and Wasserstein based transfer curriculum. For ease of comparison, the y-coordinate of the first two datasets represents $1-ACC$, where lower values indicate superior performance. Each configuration is evaluated 100 times, and the shaded areas represent the variance.}
    \label{fig:sort}
	\end{center}
 \vspace{-1cm}
\end{figure}

As shown in Fig. \ref{fig:sort}, the superior performance of COT over DOT, coupled with the comparable performance between COT utilizing w-distance and COT with true metadata, shows the effect of intermediate domain sorting. The impact of varying the number of intermediate domains differs across datasets, which is determined by the inherent characteristics of the data itself. These findings highlight the benefits of incorporating COT and the Wasserstein distributional geometric relationships.

\subsection{Adaptation Comparison Results}

A comprehensive comparison of our proposed method, W-MPOT, is conducted with several classic approaches in the realm of CDA. 
We present the comparison results in Table \ref{tab:comparison}.

In our experimental setup, we maintain a fixed number of two intermediate domains, and each experiment is repeated 100 times to ensure robustness, with the average results reported. In the bottom section of Table \ref{tab:comparison}, we showcase the outcomes of our proposed algorithm, W-MPOT, applied in three distinct scenarios: p2$\rightarrow$p1, p1$\rightarrow$p2, and p1 $+$ p2. These scenarios involve using one path to refine the other and utilizing both paths for a new refined path. Bilateral experiments are conducted by applying a mutual constraint mechanism using regularizers from both paths.

The results indicate the superior performance of W-MPOT over other methods across all three datasets, underscoring its efficacy in mitigating continuous domain shifts. Notably, W-MPOT using both path 1 and path2 shows the optimal performance, providing strong evidence for the effectiveness of the added regularization term $R_p(\cdot)$. The results of W-MPOT in other two scenarios: p2$\rightarrow$p1 and p1$\rightarrow$p2, showing similar performance, suggest that the regularization approach is robust and does not heavily rely on the specific choice of the second path. By leveraging both paths with mutual constraints, W-MPOT successfully improves the overall robustness of the model.

\begin{table}[htb]
\scriptsize
\centering
\caption{MSE or Accuracy for three datasets of different algorithms}
\setlength{\tabcolsep}{3.5mm}{
\begin{tabular}{c|ccc}
 \toprule
Method & ADNI~($\uparrow$) & Battery~($\downarrow$) & ROT MNIST~($\uparrow$) \\

 \midrule
    Source Model & 41.2 & 0.3731 & 48.1 \\
    CMA \cite{hoffman2014continuous} & 55.4 & 0.3842 & 65.3 \\
    EAML \cite{liu2020learning} & 68.3 & 0.2045 & 70.4 \\
    AGST \cite{zhou2022active} & 57.3  & 0.3534  & 76.2  \\
    Gradual ST \cite{kumar2020understanding} & 64.5  & 0.1068  &   87.9 \\
    CDOT \cite{ortiz2019cdot} & 82.6  & 0.0209  &  75.6  \\

 \midrule

\textbf{W-MPOT(p2$\rightarrow$p1)} & \underline{86.7} & {0.0199} & \underline{88.3} \\
\textbf{W-MPOT(p1$\rightarrow$p2)} & {86.5} & \underline{0.0197} & {87.2} \\
\textbf{W-MPOT(p1 + p2)} & \textbf{88.3} & \textbf{0.0185} & \textbf{89.1} \\
\bottomrule

\end{tabular}
}
\label{tab:comparison}
 \vspace{-1cm}
\end{table}

\subsection{Ablation Study}

\begin{figure}[h]
	\begin{center}
		\centerline{\includegraphics[width=0.9\columnwidth]{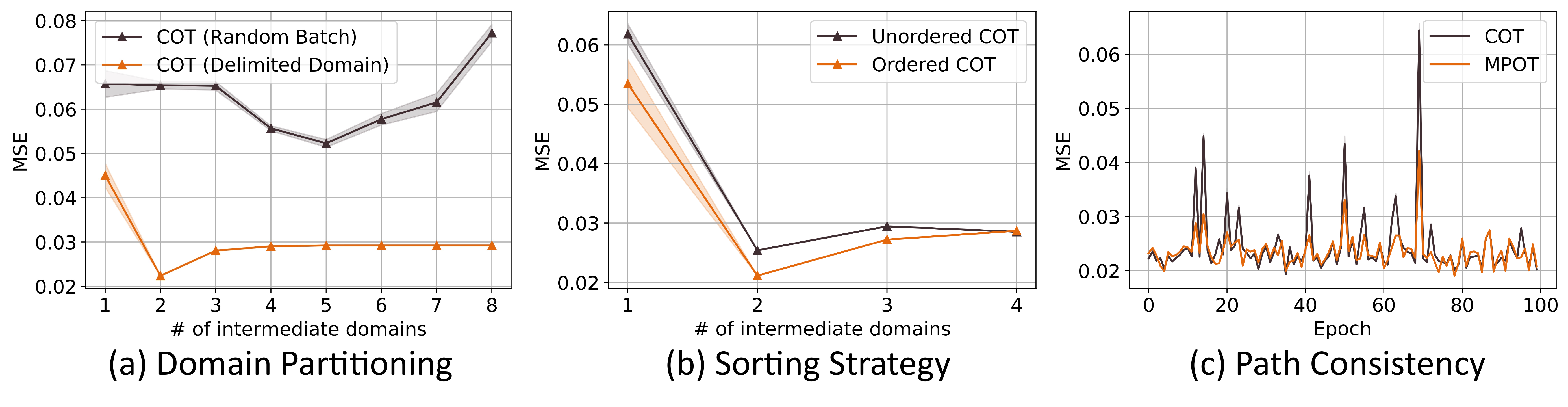}}
		\caption{\textbf{Ablation study Results.} The experiments from a to b involved domain partitioning, sorting strategy, and path consistency. For each number of intermediate domains in a and b, the experiment was randomly repeated 100 times. The solid line represents the mean MSE, while the shaded area represents the variance. The experiment in c is conducted by fixing the number of intermediate domains to 2 and randomly sampling different intermediate domains 100 times.}
		\label{fig:ablation}
	\end{center}
 \vspace{-1cm}

\end{figure}

We conducted ablation studies on the \textit{Battery Charging-discharging Capacity} dataset to investigate the effect of domain partitioning, Wasserstein-based sorting strategy, and path consistency regularization separately. Delimited domains consistently exhibit lower mean and variance of MSE compared to random batch sampling in Fig. \ref{fig:ablation}(a), indicating increased stability. This highlights the necessity of considering distinct domains for improved predictive accuracy. Comparing Unordered COT to Ordered COT reveals consistently lower MSE values in Fig. \ref{fig:ablation}(b), emphasizing the value of the Wasserstein-based transfer curriculum for superior performance in battery capacity prediction tasks. The vital role of the path consistency regularization term in achieving accurate and robust domain mappings is demonstrated by the comparison of MPOT (with $R_p(\cdot)$) and COT (without $R_p(\cdot)$) in Fig. \ref{fig:ablation}(c).





\section{Conclusion}

This research introduces the W-MPOT framework for CDA, effectively addressing substantial domain shifts and missing metadata. It comprises the Wasserstein-based Transfer Curriculum for domain ordering and MPOT for cumulative errors during adaptation. Experimental results across diverse datasets demonstrate its superior performance, highlighting the practicality and potential of these methods in handling substantial domain shift challenges. This work advances the field of CDA and offers insights into addressing domain shift effectively, especially in domains like healthcare and energy storage. Future research could focus on establishing generalization bounds and employing reinforcement learning to optimize the selection of intermediate domains, making domain adaptation a sequential decision-making process.

\section{Acknowledgement}
This work was supported in part by the Natural Science Foundation of China (Grant 62371270), Shenzhen Key Laboratory of Ubiquitous Data Enabling 

\noindent(No.ZDSYS20220527171406015), and Tsinghua Shenzhen International Graduate School Interdisciplinary Innovative Fund (JC2021006).

\bibliographystyle{splncs04}
\bibliography{mybibfile}

\appendix

\section{Optimal Transport}
This section provides the details for the Optimal transport (OT) used in the main text. OT provides a measurement of the divergence between two probability distributions by finding an optimal transportation plan that minimizes the total cost of mapping one distribution to the other. When used to measure the divergence between the source and target domain, this transportation plan specifies how mass from each point in the source distribution is moved to corresponding points in the target distribution \cite{courty2016optimal}. 
The cost of moving mass from one point to another is typically defined by a ground metric.

Formally, given two probability distributions $\mu_s$ and $\mu_t$ with $N_S$ and $N_T$ points, respectively, OT looks for a transport plan $\gamma_0$ that minimizes the total cost:

\begin{equation}
\begin{gathered}
    \gamma_0=\underset{\gamma \in \mathcal{P}}{\arg \min }\langle\boldsymbol{\gamma}, \mathbf{M}\rangle_F + \lambda \cdot \Omega(\gamma), \\
    \mathcal{P}=\left\{\gamma \in\left(\mathbb{R}^{+}\right)^{{N}_{{S}} \times {N}_{{T}}} \mid \gamma \mathbf{1}_{{N}_{{T}}}=\mu_{{S}}, \gamma^{\mathbf{T}} \mathbf{1}_{N_S}=\mu_{{T}}\right\} \\
\end{gathered}
\end{equation}

where $\mathbf{M}$ is a transportation matrix and $\Omega$ denotes the entropic regularization term $\Omega(\gamma)=\sum_{i, j} \gamma_{i, j} \log \left(\gamma_{i, j}\right)$. $\lambda > 0$ is the weight of entropic regularization. 
The optimal transport problem can be solved using various algorithms, such as the Sinkhorn algorithm \cite{cuturi2013sinkhorn}, the linear programming-based method, or the entropic regularization approach, etc. 

\section{Lemma of Generalization Bound}
We proposed the generalization bound for CDA in the main text, referencing the following lemma introduced in \cite{shen2018wasserstein}.

\noindent\begin{lemma}
    Suppose the two domains have the same true labeling function $f:\mathcal{X}\rightarrow [0,1]$. Let $\mu_S$, $\mu_T\in \mathcal{P}(\mathcal{X})$ be two probability measures on $\mathbb{R}^d$. Assume that all hypotheses in the hypothesis set $H$ satisfies the $A$-Lipschitz continuous condition for some $A$. Denote the error function of any hypothesis $h$ compared with the true labeling function with respect to distribution $\mu$ as $\epsilon_{\mu}(h,f)=\mathbb{E}_{\boldsymbol{x}\sim\mu}\left[\left|h(\boldsymbol{x}) - f(\boldsymbol{x})\right|\right]$.
    Then for any hypothesis $h \in H$, the following holds:
    $$
    \epsilon_{\mu_T}(h,f) \leq \epsilon_{\mu_S}(h,f)+2 A \cdot\mathcal{W}_{1}\left(\mu_{S}, \mu_{T}\right)+\epsilon,
    $$
    where $\epsilon$ is the combined error of the ideal hypothesis $h^*$ that minimizes the combined error 
$\epsilon_{\mu_S}(h,f) + \epsilon_{\mu_T}(h,f)$.
\end{lemma}

Let $\mu_{I_0}$, $\mu_{I_0}\in \mathcal{P}(\mathcal{X})$ be the probability measures on for domain $(D_{I_0}$, $D_{I_1})$, respectively. By simply applying this lemma to each source-target pairs $(D_S, D_{I_0})$, $(D_{I_0}, D_{I_1})$ and $(D_{I_0}, D_{T})$, the generalization bound of the transfer path $D_S \rightarrow D_{I_0} \rightarrow  D_{I_1} \rightarrow D_T$ could be derived.

\section{Detailed Solution for the Sinkhorn algorithm}
The Sinkhorn algorithm is an efficient iterative method to solve the optimal transport problem by iteratively updating a matrix of non-negative values to approach the optimal coupling \cite{cuturi2013sinkhorn}. As in \cite{ortiz2019cdot}, the MPOT problem fits into the forward-backward splitting algorithm \cite{van2017forward, bui2021bregman}, whose solution could be efficiently computed with the Sinkhorn algorithm. Therefore, Eq. (5) in the main text can be solved through the following iterative steps

\begin{equation}
\begin{gathered}
 \mathbf{M}^{[N, N+1]}_c=\alpha \mathbf{M}^{[N, N+1]}+\alpha \nabla \mathcal{J} (\gamma_c, \gamma_{p_2}), \\
\gamma_{c+1}=\operatorname{Sinkhorn}\left(\mathbf{M}^{[N, N+1]}_c, 1+\alpha \lambda, \mu_{SI_{N}}, \mu_{T}\right), \\
\end{gathered}
\label{eq:iter_gamma}
\end{equation}
where $c\in[0,\infty)$ is the index of iteration, and it will be repeated until $\gamma$ converges. $\alpha$ is the step size and $\nabla$ is the differential computation symbol. $\mathcal{J}(\cdot)$ is a combination of regularization terms as follows
\begin{equation}
\begin{split}
\mathcal{J}(\gamma, \gamma_{p_2}) = \eta_t R_t(\gamma) + \eta_p R_p(\gamma, \gamma_{p_2}) - \lambda \Omega(\gamma),
\end{split}
\end{equation}
The derivation result of $\mathcal{J}(\cdot)$ is
\begin{equation}
\begin{split}
\nabla \mathcal{J}(\gamma, \gamma_{p_2})
= \eta_t \frac{\partial R_t(\gamma)}{\partial \gamma} + \eta_p \frac{\partial R_p(\gamma, \gamma_{p_2})}{\partial \gamma} - \lambda \frac{\partial \Omega(\gamma)}{\partial \gamma},
\end{split}
\label{eq:dev_j}
\end{equation}

Since $R_p(\cdot)$ can be expressed in the form of trace with respect to $\gamma$, we then use $x_n$ to represent $\widehat{D}_{I_n}$ and take the derivative of $R_p(\cdot)$ as
\begin{equation}
\begin{split}
\frac{\partial R_p(\gamma, \gamma_{p_2})}{\partial \gamma} &=\frac{\partial \operatorname{tr}\left [ N_S^2\cdot x_n^T\cdot \gamma ^T\cdot \gamma \cdot x_n \right ]}{\partial \left ( \gamma ^T\cdot \gamma  \right )}\frac{\partial \left ( \gamma ^T\cdot \gamma  \right )}{\partial \gamma } \\
&\qquad -\frac{\partial \operatorname{tr}\left [ N_S^2\cdot x_n^T\cdot \gamma ^T\cdot \gamma_{p_2}\cdot x_n \right ]}{\partial \gamma } \\
&\qquad-\frac{\partial \operatorname{tr}\left [ N_S^2\cdot x_n^T\cdot \gamma_{p_2}^T\cdot \gamma \cdot x_n  \right ]}{\partial \gamma } \\
&=2N_S^2\left ( \gamma -\gamma_{p_2}  \right ) x_nx_n^T.
\end{split}
\label{eq:dev_rp}
\end{equation}

Similar to the derivative of $R_p(\cdot)$, the derivative of $R_t(\cdot)$ and $\Omega(\cdot)$ can be derived as follows,

\begin{equation}
\frac{\partial R_t(\gamma)}{\partial \gamma} = 2N_S^2(\gamma \cdot x_n-\gamma_{n-1} \cdot x_{n-1}) x_n^T,
\label{eq:dev_rt}
\end{equation}

\begin{equation}
\frac{\partial \Omega(\gamma)}{\partial \gamma} = \log(\gamma).
\label{eq:dev_omega}
\end{equation}

By integrating Eq. (\ref{eq:dev_rp}), Eq. (\ref{eq:dev_rt}) and Eq. (\ref{eq:dev_omega}) together into Eq. (\ref{eq:dev_j}), we can obtain the $\nabla \mathcal{J}(\cdot)$.
Then we repeat the iteration of $\gamma$ as outlined in  Eq. \ref{eq:iter_gamma}.

\section{Implementation Details}
In this section, we provide details on the implementations. We implemented optimal transport in both Wasserstein based transfer curriculum and MPOT using the POT package \footnote{https://pythonot.github.io/index.html}. All models were implemented by PyTorch and all the experiments were performed on a machine equipped with one Intel(R) Xeon(R) E5-2620 CPU and one NVIDIA TITAN V GPU.


\end{document}